\documentclass[letterpaper, 10 pt, conference]{ieeeconf}  

\IEEEoverridecommandlockouts                              

\usepackage{colortbl}
\usepackage{bm}
\usepackage{soul}
\usepackage{multirow}
\usepackage{makecell}
\usepackage{booktabs}
\usepackage{nicefrac}
\usepackage{animate}
\usepackage{wrapfig}
\usepackage{algorithm}
\usepackage[noend]{algpseudocode}
\usepackage{amssymb}
\usepackage{booktabs}
\usepackage{caption}
\usepackage{etoolbox}
\usepackage{graphicx}
\usepackage[font=small,labelfont=bf]{caption}
\usepackage[dvipsnames]{xcolor}

\usepackage{hyperref}
\hypersetup{
    colorlinks=true,
    citecolor=WildStrawberry,
    linkcolor=SeaGreen,
    urlcolor=cyan,
}
\usepackage{listings}
\usepackage{subcaption}
\usepackage{tabularx}
\usepackage{tikz}
\usetikzlibrary{fadings}
\usepackage{titlesec}
\usepackage{xcolor}
\usepackage{soul}
\usepackage{multirow}  

\usepackage{amsmath,amsfonts,bm}

\def\eqref#1{equation~\ref{#1}}

\def\1{\bm{1}}

\DeclareMathAlphabet{\mathsfit}{\encodingdefault}{\sfdefault}{m}{sl}
\SetMathAlphabet{\mathsfit}{bold}{\encodingdefault}{\sfdefault}{bx}{n}

\DeclareMathOperator*{\argmin}{arg\,min}

\definecolor{deepblue}{rgb}{0,0,0.5}
\definecolor{deepred}{rgb}{0.6,0,0}
\definecolor{magenta}{rgb}{1.0,0,1.0}
\definecolor{deepgreen}{rgb}{0,0.5,0}
\definecolor{textblue}{rgb}{.2,.2,.7}
\definecolor{textred}{rgb}{0.54,0,0}
\definecolor{textgreen}{rgb}{0,0.43,0}
\definecolor{es-blue}{rgb}{0.1372,0.666,1}
\definecolor{stefan}{rgb}{0.,0.33.,0.0}
\definecolor{hanzhi}{rgb}{0.08,0.33,0.6}
\definecolor{yao}{rgb}{1.0,0.5,0.0}
\definecolor{allcolor}{rgb}{1.0,0.53,0.0}
\definecolor{author}{rgb}{0.2,0.1,0.6}
\definecolor{seoul}{rgb}{0.0,0.71,0.57}
\definecolor{lightblue}{rgb}{0.7, 0.9, 0.98}
\definecolor{PeachPuff}{rgb}{1 0.85 0.73}
\definecolor{powderblue}{rgb}{0.753, 0.847, 0.902}
\definecolor{lightsteelblue}{rgb}{0.690, 0.769, 0.871}
\definecolor{darkseagreen}{rgb}{0.561, 0.737, 0.561}

\newcolumntype{Y}{>{\centering\arraybackslash}X}

\usepackage[capitalise, nameinlink]{cleveref}
\makeatletter
\let\NAT@parse\undefined
\makeatother
\usepackage[numbers,sort&compress]{natbib}

\newcommand{\authorhref}[3][author]{\href{#2}{\color{#1}{#3}}}

\newcommand{\methodname}{\textsc{GOPLA}}

\begin{document}

\title{\LARGE \bf
\textcolor{darkseagreen}{\methodname}: \textcolor{darkseagreen}{G}eneralizable \textcolor{darkseagreen}{O}bject \textcolor{darkseagreen}{P}lacement \textcolor{darkseagreen}{L}earning via Synthetic \textcolor{darkseagreen}{A}ugmentation of Human Arrangement
}

\author{
  \authorhref{https://www.linkedin.com/in/yao-zhong-72051223a/}{Yao Zhong}$^{*, 1}$    \quad
  \authorhref{https://hanzhic.github.io/}{Hanzhi Chen}$^{*, 1, 2}$     \quad  
  \authorhref{https://simon-schaefer.github.io/}{Simon Schaefer}$^{1, 2}$   \quad  
  \authorhref{https://dipan-zhang.github.io/}{Anran Zhang}$^{1, 2}$  \quad    
  \authorhref{https://scholar.google.ch/citations?user=SmGQ48gAAAAJ}{Stefan Leutenegger}$^{2}$ \quad \\
  $^*{\text{ Equal Contribution }}$ $^1\text{ Technical University of Munich }$ $^2\text{ ETH Zurich  }$
}


\renewcommand{\baselinestretch}{0.97}

\twocolumn[{
\renewcommand\twocolumn[1][]{#1}%
\maketitle
\begin{center}
    \centering
    \vspace{-30pt}
    \includegraphics[width=1.0\textwidth]{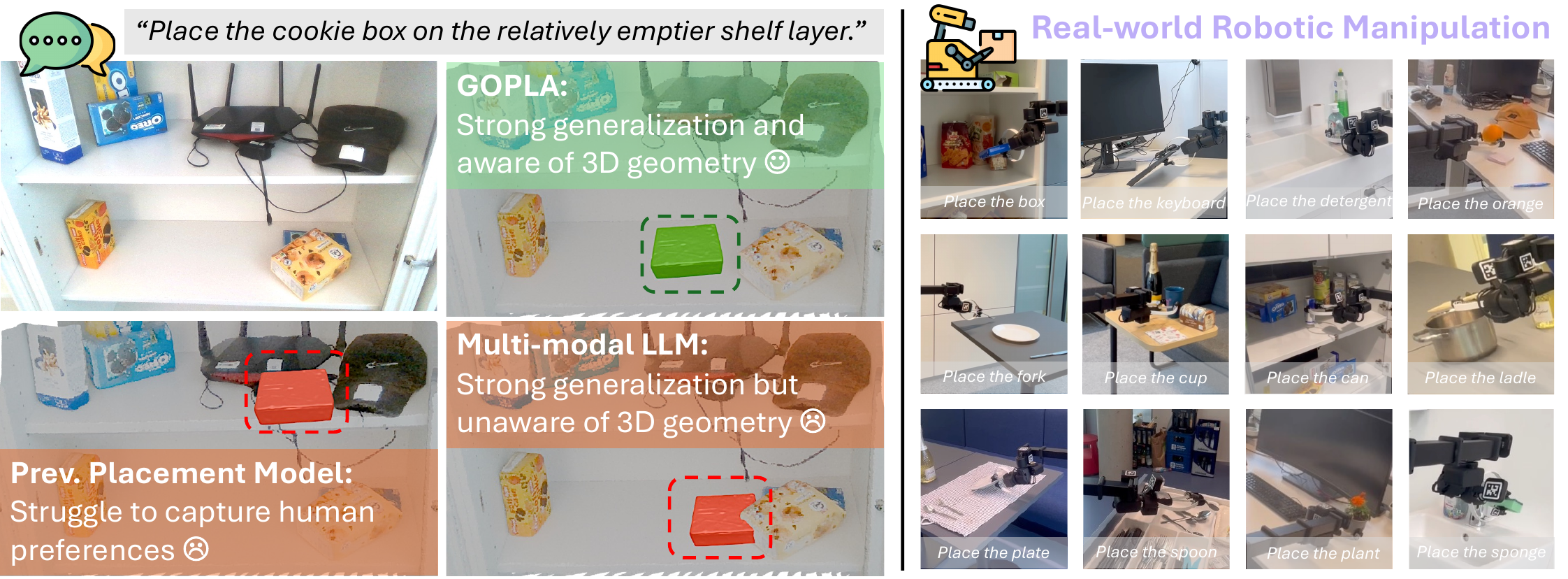}   
    \vspace{-10pt}
    \captionsetup{type=figure}\caption{{We present \textbf{\textcolor{darkseagreen}{\methodname}}, a framework that uses synthetic augmentation of real-world human arrangement scenes to achieve generalizable object placement that respects both human preferences and 3D spatial constraints. }}
    \label{fig:teaser_fig}
    \vspace{-7pt}
\end{center}
}] 


\thispagestyle{empty}
\pagestyle{empty}



\begin{abstract}
Robots are expected to serve as intelligent assistants, helping humans with everyday household organization. A central challenge in this setting is the task of object placement, which requires reasoning about both semantic preferences (e.g., common-sense object relations) and geometric feasibility (e.g., collision avoidance). We present \methodname, a hierarchical framework that learns generalizable object placement from augmented human demonstrations. A multi-modal large language model translates human instructions and visual inputs into structured plans that specify pairwise object relationships. These plans are then converted into 3D affordance maps with geometric common sense by a spatial mapper, while a diffusion-based planner generates placement poses guided by test-time costs, considering multi-plan distributions and collision avoidance. To overcome data scarcity, we introduce a scalable pipeline that expands human placement demonstrations into diverse synthetic training data. Extensive experiments show that our approach improves placement success rates by 30.04 percentage points over the runner-up, evaluated on positioning accuracy and physical plausibility, demonstrating strong generalization across a wide range of real-world robotic placement scenarios.
\end{abstract}
\section{Introduction}
\label{sec:intro}

    


Humans effortlessly place objects in various daily scenes, adhering to common sense and physical constraints. For example, when configuring a workspace for a left-handed user, the intuitive spatial arrangement is to position the pen to the left of the notebook, ensuring both accessibility and a collision-free environment. Future robots will serve as physical assistants, requiring similar capabilities to arrange environments according to user preferences. However, learning object placement remains challenging, as it demands both semantic understanding to recognize contextual appropriateness and geometric awareness to satisfy spatial constraints. Robots must also process high-dimensional task inputs, including 3D scene representations and language-guided preferences, to make informed decisions. Mastering this skill is crucial for developing intelligent agents capable of autonomous household organization.

To address this challenge, a line of work designs placement models by collecting scene-specific or task-specific datasets and training a model to predict plausible placement configurations \cite{kapelyukh2022my, ramrakhya2024seeing, jia2024cluttergen, mo2021o2oafford}. However, scaling up such expensive training data with both visual common sense and privileged geometric labels poses an extreme challenge. As a result, these models can only capture one limited aspect, either visual common sense \cite{kapelyukh2022my, ramrakhya2024seeing} or physical plausibility \cite{jia2024cluttergen, mo2021o2oafford}. They also typically require fixed-format text inputs and struggle to interpret the nuanced human preferences expressed in free-form instructions. Another line of work adopts multi-modal large language models (MLLMs) with strong reasoning ability.
The MLLMs are tasked to provide answers about plausible placement locations \cite{yuan2024robopoint, nasiriany2024pivot, liu2024moka}. Nonetheless, they struggle with handling complex scenarios where spatial constraints (e.g., collision avoidance) have to be satisfied, since both the 3D spatial information from depth and fine-grained objects' geometric properties are fully overlooked \cite{kamath2023s, yamada2023evaluating, shiri2024empirical}. 

In this work, we seek to answer one fundamental question: \textit{how to design a generalizable placement model despite the combinatorial explosion of scenes, objects, and user instructions?}
Our key insight is that rather than training a highly data-intensive MLLM \cite{yuan2024robopoint, deitke2024molmo, song2025robospatial}, we leverage the strong reasoning capability of a pre-trained MLLM to transform an \textbf{open-ended condition} (e.g., \textit{“I am left-handed, help me place the pen.”}) into a \textbf{closed set of structured plans} (e.g.,  \textit{left of the notebook}, \textit{right front of the cellphone}, etc.). 
These structured plans capture pairwise object relationships and can be efficiently handled by specialist spatial reasoning models requiring structured inputs.

We propose a hierarchical placement model.
First, a high-level MLLM captures human preferences and generates multiple structured plans in parallel from vision-language inputs.
Next, a mid-level spatial mapper converts these plans into 3D affordance maps, embedding geometric common sense. 
Finally, a low-level diffusion-based planner synthesizes placement poses, guided by test-time cost to combine all proposed pairwise inter-object relationships and enforce physical plausibility. 
Using these cost functions, the planner adapts to unseen plan combinations and novel scene contexts, leading to substantial generalization.
Though training lower-level models is much cheaper than MLLMs, their training data must capture spatial relationships between objects that align with human common sense. For instance, a mouse is typically placed approximately 10 cm from a keyboard rather than 1 meter away. Collecting such training data remains labor-intensive, which brings us to our second question: \textit{how to scale training data that reflects both human common sense and reasonable spatial relationships?} To address this, we develop an automated data generator to scale real-world demonstrations into the synthetic domain. We represent human placement scenes as scene graphs, diversify arrangements and object categories based on functional similarity, and refine the layouts to generate realistic training data.

We validate our approach through extensive experiments across diverse synthetic and real-world environments. Our method outperforms several strong baselines~\cite{yuan2024robopoint, ramrakhya2024seeing, gpt4o, liu2023llava}, achieving a 30.04 percentage points average improvement in success rate. 
Success is evaluated based on alignment with user instructions and collision-free physical plausibility. Remarkably, despite being trained on low-cost synthetic data augmented from a few human demonstrations, our model exhibits strong generalization to real-world human-suited environments. We demonstrate the versatility of our model by deploying it on a real robot and conducting several household arrangement tasks. 

To summarize, our key contributions are: (1) A hierarchical placement model aware of human preferences and 3D spatial constraints, ensuring semantic relevance and physical plausibility. (2) A scalable data generation pipeline that uses few human demonstrations to create diverse synthetic training data. (3) Extensive experiments in both synthetic and real-world environments to showcase the effectiveness, generalization, and versatility of our framework, \methodname. 
\section{Related Works}
\label{sec:related}

\vspace{-2pt}
\noindent \textbf{Multi-modal LLMs for Robotic Manipulation.} 
Recent works \cite{kim24openvla, progprompt, nasiriany2024pivot, yuan2024robopoint, liu2024moka, song2025robospatial, cheng2024spatialrgpt} have shown the efficacy of MLLMs in robotic tasks. RoboPoint \cite{yuan2024robopoint} tunes MLLMs to predict keypoint affordance for robotic applications, which is used as a simplified action representation. Such a representation also shows effectiveness through visual prompting as in \cite{nasiriany2024pivot, liu2024moka, deitke2024molmo, zhao2025anyplace}. However, MLLMs' limited spatial reasoning capability makes it challenging to handle spatial constraints induced by the geometric variances of the objects and the scenes \cite{song2025robospatial, cheng2024spatialrgpt}. In contrast, we employ a MLLM to transform open-ended preferences into structured representations, which are then processed by low-level spatial reasoning models, enabling the generation of geometry-aware actions.

\noindent \textbf{Object Placement Affordance Learning.}
Affordance describes object-centric functionality, focusing on \textit{where} and \textit{how} an agent can interact with an object \cite{li2022gendexgrasp, mo2021o2oafford, chen2025vidbot}. 
In the domain of placement tasks, this term refers to a \textit{plausible pose} to place the object given scene context so that several criteria could be satisfied, e.g., aesthetics, functionality, plausibility, etc. 
Recent works \cite{kapelyukh2022my, ramrakhya2024seeing, yuan2024robopoint} have explored learning common sense for placement using web data. However, geometric variances of the scenes are often overlooked, resulting in physically implausible configurations causing collisions with other scene objects. Another line of works~\cite{mo2021o2oafford, jia2024cluttergen, weng2023towards} explored geometry-aware placement planning, while the predictions fail to capture human preferences. In contrast, our hierarchical placement model can not only capture user preferences but also respect 3D spatial constraints.
Concurrent with our work, AnyPlace~\cite{zhao2025anyplace} designs a two-stage model for placement, but it is tailored to more controlled in-lab setups with limited scene complexity. In contrast, our model generalizes robustly to in-the-wild environments demanding richer semantic and geometric reasoning.


\noindent \textbf{Object Placement Data Generation.} 
Existing dataset generation methods for placement tasks face several key limitations: missing 3D labels~\cite{kapelyukh2022my, ramrakhya2024seeing}, insufficient semantics~\cite{jia2024cluttergen, mo2021o2oafford}, simulator-induced errors~\cite{yuan2024robopoint, mo2021o2oafford}, and limited diversity~\cite{zhao2025anyplace, simeonov2023rpdiff}.
In contrast, inspired by MimicGen~\cite{mandlekar2023mimicgen}, our data generator leverages a few real-world scenes as demonstrations to produce diverse synthetic data at scale, enriched with both privileged geometric labels and human common sense. 
\section{Method}

\subsection{Problem Formulation} 
\vspace{-4pt}
The problem we aim to address is predicting placement poses that adhere to physical constraints and human preferences based on observed scenes and language instructions. Our placement model is formulated as $\bm{T}_\text{CO} = \pi \left( (  \bm{I} ,  \bm{D}  ), \bm{O}, l \right)$, where  \( (  \bm{I} ,  \bm{D} )\) is an RGB-D image pair, $l$ is a free-form language instruction, and $\bm{O}$ are the sampled surface points from the object mesh to be placed (obtained via single-view 3D generation \cite{long2023wonder3d, xu2024instantmesh} or 3D reconstruction \cite{dust3r_cvpr24, mast3r_arxiv24}). $\bm{T}_\text{CO}$ is the predicted placement pose relative to the camera frame, and we denote it as $\bm{T}$ for simplicity.

\begin{figure*}[t]
    \vspace{0.1 in}
    \centering
    \includegraphics[width=\linewidth]{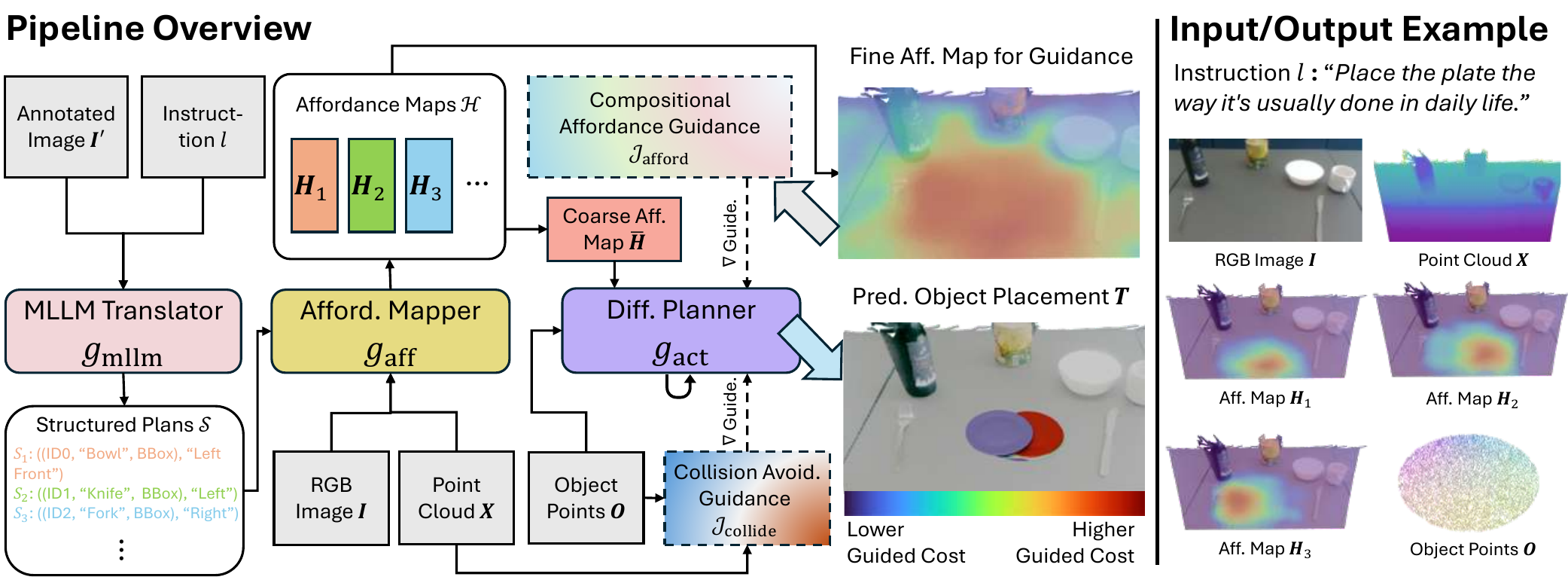}
    
    \vspace{-0.08 in}
    \captionsetup{type=figure}\caption{Overview of our proposed placement model. A hierarchical model is used to infer the pose of an object to be placed, capturing both the user preference and the physical plausibility.}
    \label{fig:pipeline}
    \vspace{-20pt}
\end{figure*}


\subsection{Hierarchical Object Placement Learning}
\vspace{-4pt}
\label{pipeline}
We design our model based on two key considerations: (1) It should capture user preferences expressed through free-form language instructions. (2) It should account for 3D spatial constraints imposed by the scenes. As in Fig.~\ref{fig:pipeline}, a MLLM $g_{\mathrm{mllm}}$ translates open-ended observations into structured plans, and two spatial reasoning models, $g_{\mathrm{aff}}$ and $g_{\mathrm{act}}$, are conditioned on these plans to translate them into 3D affordance maps and to generate placement poses with geometric awareness.

\noindent\textbf{High-level MLLM Translator.} In the initial stage of the pipeline, the translator maps free-form vision-language inputs into structured plans, enabling downstream spatial reasoning models to effectively parse them.
An open-set detector \cite{liu2023grounding} is used to get anchor object proposals $\mathcal{P}=\{(n_i, c_i, \bm{b}_{i})\}_{i=1}^{N_{\text{p}}}$ from the color image $\bm{I}$, where $n_i$ is the object's ID, $c_i$ is the class name, and $\bm{b}_{i} \in \mathbb{R}^{4}$ is the bounding box. To apply the visual prompting strategy \cite{nasiriany2024pivot, yang2023setofmark}, we annotate the image $\bm{I}$ with the bounding boxes and corresponding IDs from $\mathcal{P}$, producing an annotated image $\bm{I}^\prime$. A pretrained MLLM \cite{gpt4o} $g_{\mathrm{mllm}}$  is used to compute a set of structured plans with $\mathcal{S} = g_{\mathrm{mllm}}(\bm{I}^\prime, l, q_{\mathrm{mllm}})$, where $q_{\mathrm{mllm}}$ is the prompt. For structured plans $\mathcal{S}=\{((n_i, c_i, \bm{b}_{i}), d_i)\}_{i=1}^{N_{\text{s}}}$, $d_i$ is a direction text like \textit{right}, \textit{left front} from a predefined relation set $\mathcal{D}$. Such a plan, e.g., left of ID 1 (\textit{knife}), encodes a pairwise relationship for a primitive placement plan. 

\noindent \textbf{Mid-level Spatial Mapper.} This module serves as a spatial grounding bridge.
A spatial mapper $g_{\mathrm{aff}}$ translates each structured plan into 3D space, producing a per-point affordance activation map that can be explicitly composed.
We acquire the scene point cloud $\bm{X} \in \mathbb{R}^{N_{\text{x}}}$ by back-projecting the depth image $\bm{D}$ using camera intrinsics $\bm{K}$. 
We crop the object point cloud $\bm{X}_i$ from the scene cloud $\bm{X}$ using $\bm{b}_i$, and compute its position $\bm{p}_i \in \mathbb{R}^3$ as the component-wise median of $\bm{X}_i$.
We extract the geometric feature of $\bm{X}$ using a PointNet++ \cite{qi2017pointnetplusplus} encoder $f_{\mathrm{geo}}$: $\bm{z}_{\text{g}}  = f_{\mathrm{geo}}(\bm{X})$, and the semantic feature of $\bm{I}$ using an extractor \cite{ramrakhya2024seeing} $f_{\mathrm{sem}}$: $\bm{z}_{\text{s}}  = f_{\mathrm{sem}}(\bm{I}, c_i)$. Next, we propagate the color-aligned semantic feature $\bm{z}_{\text{s}} $ to each point from $\bm{X}$ to acquire $\bm{z}_{\text{s}}^\prime$ to establish the global scene context feature $\bm{z}_{\text{t}} = [\bm{z}_{\text{g}}, \bm{z}_{\text{s}}^\prime]$.
To attend to the structured plan $((n_i, c_i, \bm{b}_{i}), d_i)$, we employ two Perceiver \cite{perceiver} modules, $f_\mathrm{perc}^1$ and $f_\mathrm{perc}^2$, to decode the affordance value ${\bm{H}_i} \in \mathbb{R}^{N_{\text{x}}}$: 
\begin{equation}
\begin{split}
\bm{u}_i &= f_\mathrm{perc}^1\!\left(
    \bm{z}_{\mathrm{t}},\ 
    f_\mathrm{proj}^2\!\left(
        \left[
            f_\mathrm{proj}^1(\bm{p}_i),\ 
            f_\mathrm{text}(c_i)
        \right]
    \right)
\right), \\[4pt]
\bm{H}_i &= f_\mathrm{perc}^2\!\left(
    \bm{u}_i,\ 
    f_\mathrm{text}(d_i)
\right). \label{eqn:affordance_decode}
\end{split}
\end{equation}
Here, $\bm{u}_i$ is an intermediate feature generated by $f_\mathrm{proj}^1$. $f_\mathrm{proj}^1$ and $f_\mathrm{proj}^2$ are blocks with a MLP and a transformer encoder \cite{vaswani2017attention}, and $f_\mathrm{text}$ is a frozen CLIP text encoder \cite{CLIP}.  
We denote this process as $\mathcal{H}=\{\bm{H}_i\}_{i=1}^{N_{\text{s}}}, \bm{H}_i = g_{\mathrm{aff}}(\bm{X}, \bm{I}, (\bm{p}_i, c_i, d_{i}))$.

\noindent\textbf{Low-level Diffusion Planner.} This module operates immediately after the spatial mapper to produce the final placement pose. The aim of the planner $g_{\mathrm{act}}$ is: 1) Compose all translated plans; 2) Avoid collisions for better physical plausibility.
We model this process as a conditional diffusion denoising process \cite{diffusion}, allowing us to use test-time cost guidance during sampling to improve generalization to unseen scenarios \cite{janner2022diffuser}.
In the forward process, we sample the pose $\bm{T}^0$ from the distribution $q(\bm{T})$ and add Gaussian noise iteratively for $K$ steps:
\vspace{-1 ex}
\begin{equation}
\begin{split}
    q(\bm{T}^{1:K} | \bm{T}^0) &= \prod_{k=1}^K q (\bm{T}^k | \bm{T}^{k-1}), \\
    q(\bm{T}^k | \bm{T}^{k-1}) &= \mathcal{N}(\bm{T}^k; \sqrt{1 - \beta_k}\bm{T}^{k-1}, \beta_k \bf{I}),
\end{split}
\end{equation}

\label{eqn:forward}
\vspace{-0.5 ex}\noindent where $\beta_k$ is the variance at step $k$. In the reverse process, a network is learned to model the denoising process and gradually acquires the plausible transformation under condition $\bm{z}_\text{c}$ to be explained:
\vspace{-0.5 ex}
\begin{equation}
    p_{\phi}(\bm{T}^{k-1} | \bm{T}^k, \bm{z}_\text{c})= \mathcal{N}(\bm{T}^{k-1}; \bm{\mu}_k, \bm{\Sigma}_k),
\end{equation}

\label{eqn:reverse}

\vspace{-0.5 ex}\noindent where $\bm{\Sigma}_k$ is a variance matrix from a pre-defined scheduler. $\bm{\mu}_k$ is predicted by a learned model. 

We compute a coarse affordance map $\bar{\bm{H}}$ by applying k-means to $\mathcal{H}$ and measuring scene cloud $\bm{X}$ distance to the top-$K$ cluster centers, yielding a distance-based activation map. A PointNet++ \cite{qi2017pointnetplusplus} $f_\text{obj}$ is used to extract the object feature $\bm{z}_\text{obj} = f_\mathrm{obj}(\bm{O})$. We sample $k_\text{a}$ points based on the activation of $\bar{\bm{H}}$, and concatenate their geometric features to form $\bm{z}_\text{aff}$. We acquire the conditional feature $\bm{z}_\text{c} = [\bm{z}_\text{obj}, \bm{z}_\text{aff}]$.  
To incorporate the spatial awareness, we compute a weighted sum of scene points' affordance values \(a^k_j\), using their inverse distances \(w^k_j\) to the translation of \(\bm{T}^k\), resulting in spatial feature \(a^k = \nicefrac{\sum_j w^k_j a^k_j}{\sum_j w^k_j}\).
Instead of using noise prediction, $g_{\mathrm{act}}$ infers the unnoised pose $\bar{\bm{T}}^0$ to compute $\bm{\mu}_k$ in each step $k$: $\bar{\bm{T}}^0=g_{\mathrm{act}}((\bm{T}^k, a^k), \bm{z}_\text{c}, k)$.
The denoising network in $g_{\mathrm{act}}$ is adapted from \cite{rempeluo2023tracepace}.
To reinforce finer affordance conditioning of $\mathcal{H}$ with an arbitrary number of plans, we formulate the multi-plan composition as a cost function $\mathcal{J}_\text{afford}$ to steer the denoising process using test-time guidance \cite{janner2022diffuser}. The physical plausibility is improved via a scene collision cost $\mathcal{J}_\text{collide}$.
\begin{equation}
\begin{split}
\mathcal{J}_{\text{afford}} &=  \sum_{i} \min_{\bm{x} \in \bm{X}_\text{h}}  
\|\bar{\bm{T}}^0 \bm{O}_i - \bm{x}\|^2_2, \\
\mathcal{J}_{\text{collide}} &=  \sum_{i} - \min\left(
\operatorname{TSDF}(\bm{X}_\text{c})[\bar{\bm{T}}^0 \bm{O}_i],\ 0
\right).
\end{split}
\label{eqn:guidance}
\end{equation}
Here, $\mathcal{J}_{\text{afford}}$ is a quadratic point-set distance that measures alignment between the transformed object points $\bar{\bm{T}}^0 \bm{O}_i$ and high-affordance regions $\bm{X}_\text{h} \subseteq \bm{X}$. $\mathcal{J}_{\text{collide}}$ is a non-positve penetration penalty term. We query the signed distance value of each transformed points $\bar{\bm{T}}^0 \bm{O}_i$ using the TSDF map acquired by voxelizing $\bm{X}_\text{c} \subseteq \bm{X}$, which is the point cloud including all detected objects from $\mathcal{P}$. $\mathcal{J}_{\text{collide}}$ penalizes object points inside occupied space, with its gradient pushing the object out of collision.
The final test-time guidance function $\mathcal{J}$ is defined as $\mathcal{J} = \lambda_{\text{a}} \mathcal{J}_{\text{afford}} + 
\lambda_{\text{c}} \mathcal{J}_{\text{collide}}$, where $\lambda_{\text{a}}$ and $\lambda_{\text{c}}$ control the guidance strength. 
At each denoising step, the gradient of $\mathcal{J}$ w.r.t.\ $\bm{T}^{k}$, denoted as $\nabla_{\bm{T}^k}\mathcal{J}$,  will adjust the unnoised predictions $\bar{\bm{T}}^{0}$ as $\bm{T}^{0}$ following the \textit{reconstruction guidance} \cite{ho2022video}: $ \bm{T}^{0}= \bar{\bm{T}}^{0} \boxplus (-\bm{\Sigma}_k\nabla_{\bm{T}^k}\mathcal{J})$.
$\bm{\Sigma}_k$ is the variance schedule at step $k$, controlling the scale of the gradient update. 
The final guided cost $\mathcal{J}$ provides an intuitive metric to measure the prediction quality, which is an advantage over previous methods, e.g., \cite{ramrakhya2024seeing, yuan2024robopoint}. To balance the strength of each guidance term, a hyperparameter search is performed on a validation set when determining $\lambda_{\text{a}}$ and $\lambda_{\text{c}}$.

\vspace{-4pt}
\subsection{Scalable Placement Data Generation}
\label{sec:datagen}
 \vspace{-4pt}
Our goal is to generate a large synthetic placement dataset from a few human demonstrations provided by \cite{xu2022toscene} to train the two lower-level models, $g_{\mathrm{aff}}$ and $g_{\mathrm{act}}$.

\begin{figure}
\vspace{0.08 in}
\centering
\includegraphics[width=\linewidth]{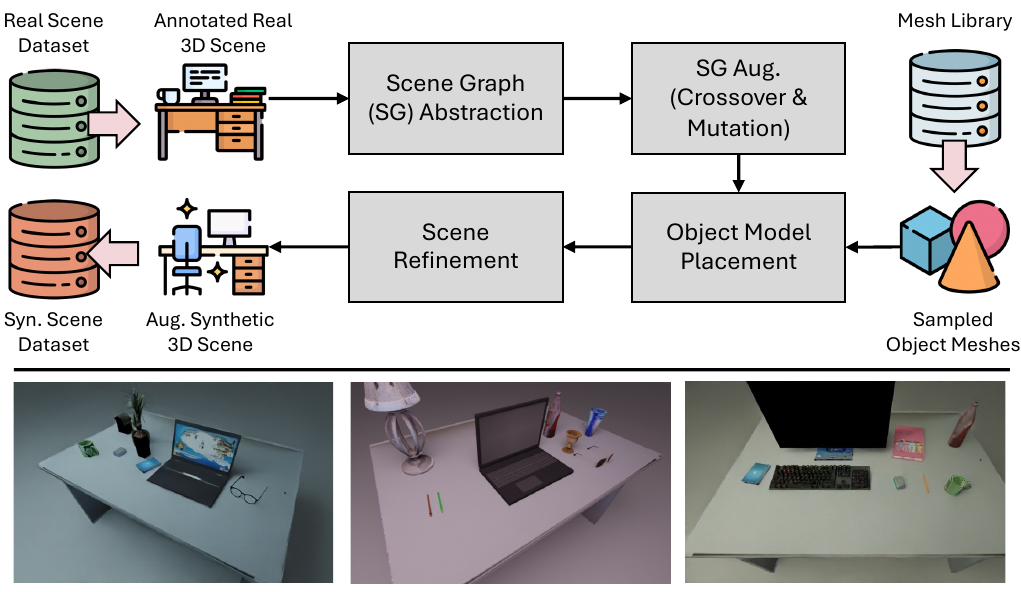}
\vspace{-0.2 in}
\captionsetup{type=figure}\caption{\small{Overview of our data generation pipeline and the generated data samples augmented from the real scene.}}
\label{fig:pipeline_datagen}
\vspace{-0.32 in}
\end{figure}

\noindent \textbf{Scene Graph Abstraction.} We first abstract each 3D scene as a scene graph \cite{Johnson_2015_CVPR_scene_graph}. For a scene $\mathcal{G} = (\mathcal{V}, \mathcal{E})$, $\mathcal{V} = \{v_1, v_2, \dots, v_N\}$ is the set of nodes representing objects in the scene. Each node $v_i \in \mathcal{V}$ is parameterized by $v_i = (c_i, \bm{p}_i)$, where $c_i$ is the object category and $p_i$ is the object centroid.
$\mathcal{E}  \subseteq \mathcal{V}  \times \mathcal{V}$ is the set of edges representing pairwise relationships between objects. Each edge $(v_i, v_j) \in \mathcal{E}$ encodes the precise spatial relationship between two objects, which can be heuristically mapped to a relation in the predefined set $\mathcal{D}$ for training.
We select the root node \( v_\text{root} \) by choosing the object closest to the scene's geometric center \( \bm{p}_\text{scene} \). \( v_\text{root} \) is defined as: 
\begin{equation}
\begin{aligned}
v_\text{root} = \argmin_{v_i \in \mathcal{V}} \| \bm{p}_i - \bm{p}_\text{scene} \|_2.
\end{aligned}
\end{equation}
Starting from the root $v_\text{root}$, we build the scene graph by connecting the nearest remaining object to its closest node in the current graph. \( \mathcal{V}_\text{s}\) denotes the set of objects added to the graph, and \( \mathcal{V}_\text{u} \) represents the set of unassigned objects. At each iteration, the next node \( v_i \) is selected as:  
\begin{equation}
\begin{aligned}
v_i = \argmin_{v_k \in \mathcal{V}_\text{u}} \min_{v_j \in \mathcal{V}_\text{s}} \|\bm{p}_k - \bm{p}_j \|_2.
\end{aligned}
\end{equation}
The selected object \( v_i \) is then connected to its nearest node \( v_j \) in the current scene graph. This process is looped until all objects are added to the graph.

\noindent \textbf{Augmentation with Crossover and Mutation.} 
To enhance scene graph diversity, we adopt the crossover and mutation operations \cite{holland-genetic-algorithms-1992}. For the crossover operation, consider two scene graphs  $\mathcal{G}_1$ and $\mathcal{G}_2$, each with a selected edge, i.e., $e_i = (v_i, v_j)$ in $\mathcal{G}_1$ and $e_m = (v_m, v_n)$ in $\mathcal{G}_2$. If the categories of the connected nodes match, i.e., $c_i = c_m$ and $c_j = c_n$, we swap the descendant subgraphs attached to $e_i$ and $e_m$ with probability $p_\text{c}$.
For the mutation operation, we define a function similarity score $s_\text{f}$ to measure the function similarity between objects based on CLIP text embeddings \cite{CLIP}: 
\begin{equation}
    \begin{aligned}
        \bm{h}_i = f_\mathrm{text}(c_i), \quad
        \bm{h}_k = f_\mathrm{text}(c_k),\quad
        s_\text{f} = \frac{\bm{h}_i \cdot \bm{h}_k}{\|\bm{h}_i\|\|\bm{h}_k\|}.
    \end{aligned}
\end{equation}
To illustrate with $\mathcal{G}_1$, if the object $v_i$ in $\mathcal{G}_1$ is \textit{functionally similar} to $v_k$ from a large object mesh library \cite{modelnet, shapenet}, i.e., $s_\text{f} > \tau_\text{p} $, we replace $v_i$ with $v_k$ with a probability $p_\text{m}$ (e.g., replacing a mug with a bottle). $\tau_\text{p}$ is a predefined threshold.
\begin{table*}[t]
    \vspace{0.1in}
    \centering
    \small
    \setlength{\tabcolsep}{10.5pt}
    \fontsize{7}{8}\selectfont
    \begin{tabular}{@{}r|ccc | ccc |ccc@{}}
        \toprule
         & \multicolumn{3}{c}{\textbf{Syn.\ Easy}} & \multicolumn{3}{c}{\textbf{Syn.\ Hard}} & \multicolumn{3}{c}{{\textbf{Real}}}\\
          \cmidrule(lr){2-4} \cmidrule(lr){5-7} \cmidrule(lr){8-10}
         
        \textbf{Variant} & \textbf{PA} $\uparrow$ & \textbf{PP} $\uparrow$ & \textbf{SR} $\uparrow$ & \textbf{PA} $\uparrow$ & \textbf{PP} $\uparrow$ & \textbf{SR} $\uparrow$ & \textbf{PA}  $\uparrow$ & \textbf{PP} $\uparrow$ & \textbf{SR} $\uparrow$ \\
        \cmidrule(lr){1-1}\cmidrule(lr){2-4} \cmidrule(lr){5-7} \cmidrule(lr){8-10}
        LLaVA-OneVision-0.5B \cite{li2024llavaonevisioneasyvisualtask}         & 34.95  &  40.66   & 25.93 & 37.46  &  24.92   & 9.68 & 33.37  &  32.42   & 9.52   \\
        LLaVA-OneVision-7B \cite{li2024llavaonevisioneasyvisualtask}           & 44.12  &  53.86   & 28.51 & 59.24  &  30.91   & 23.99 & 60.78  &  36.40   & 22.25   \\
        GPT-4o \cite{gpt4o}    & 65.05  &  79.78   & 55.16 & 66.95  &  52.17   & 27.53 &70.85  &  44.17   & 34.08    \\
        RoboPoint \cite{yuan2024robopoint}             & 63.48  &  25.90   & 16.77 & 54.67  &  14.92   & 9.57 & 58.91  &  18.35   & 12.31  \\
        SP-Policy \cite{ramrakhya2024seeing}             & 21.71  &  37.78   & 5.98   & 33.09  &  11.62   & 2.56 & 11.02  &  28.34   & 5.30\\
        \textbf{Ours}                                    & \textbf{83.95}  & \textbf{86.81}    & \textbf{73.41} & \textbf{71.33}  & \textbf{81.52}    & \textbf{63.16} & \textbf{80.43}  & \textbf{86.81}    & \textbf{70.33}\\
        \bottomrule
    \end{tabular}
    \vspace{-0.05 in}
    \captionsetup[table]{font=small}\caption{\small{Placement evaluation on synthetic and real-word splits.  \textbf{PA} denotes \underline{P}lacement \underline{A}ccuracy. \textbf{PP} deonotes \underline{P}hysical \underline{P}lausibility. \textbf{SR} denotes \underline{S}uccess \underline{R}ate. $\uparrow$: higher is better.}}
    \label{tab:benchmark}.  
    \vspace{-0.4 in}

\end{table*}

\noindent \textbf{Object Placement and Scene Refinement.}  
Referring to each node's category, we randomly sample a mesh model from the aforementioned large object library \cite{modelnet, shapenet}, and assign a texture either from the mesh's original assets or generated via \cite{chen2023text2tex}, depending on availability. 
We refine placements by first choosing the receptacle whose size most closely aligns with that of the original scene.
After retrieving object models, we perform collision checking to identify overlapping instances. For each instance in collision, we fix the poses of all others and compute their combined TSDF volumes. We optimize the pose of the colliding object by maximizing the TSDF values at its surface points while applying a regularization term to keep the pose close to its initial value for geometric consistency. Empirically, 10 gradient steps sufficed for good plausibility, 
and we remove any objects that remain in collision afterward.
For each scene, we randomly drop an object as the one to be placed, render a RGB-D image of the scene using \texttt{BlenderProc} \cite{blenderproc}, and randomly sample anchor objects to obtain the structured plans $\mathcal{S}=\{((n_i, c_i, \bm{b}_{i}), d_i)\}_{i=1}^{N_{\text{s}}}$. These inputs condition our lower-level models detailed in Sec.~\ref{pipeline}, which are supervised using the ground-truth pose $\hat{\bm{T}}$ and the affordance map $\hat{\bm{H}}$ computed using the dropped object's pose. Fig.~\ref{fig:pipeline_datagen} exemplifies the generated scenes. 


\vspace{-4pt} \subsection{Placement Model Training}
\vspace{-4pt}
We train the lower-level models using the dataset from Sec.~\ref{sec:datagen}, with $\mathcal{L}_{\mathrm{aff}}$ for the mid-level affordance mapper $g_{\mathrm{aff}}$ and $\mathcal{L}_{\mathrm{act}}$ for the low-level diffusion planner $g_{\mathrm{act}}$.
\begin{equation}
\begin{split}
\mathcal{L}_\text{aff} &=  \operatorname{BCE}( \hat{\mathbf{H}}, \mathbf{H}), \\
\mathcal{L}_\text{act} &= \mathbb{E}_{k \sim \mathcal{U}\{1,...,K\},\ \bm{T}^k \sim q(\bm{T}^k|\hat{\bm{T}})}
\left[  \left\Vert\hat{\bm{T}} - \bar{\bm{T}}^{0} \right\Vert_2^2\right],
\end{split}
\label{eqn:loss}
\end{equation}
where $\operatorname{BCE}$ is the binary cross-entropy loss. $\mathbf{H}$ is predicted by $g_{\mathrm{aff}}$, $k$ indexes the diffusion step, $\bar{\bm{T}}^{0}$ is predicted by $g_{\mathrm{act}}$ given $\bm{T}^k$. 

\subsection{Implementation Details} 
 \vspace{-4pt}
\noindent\textbf{Training Protocol.} The spatial mapper is trained using Adam~\cite{kingma2014adam} (learning rate 1e-4, batch size 4) with a StepLR scheduler decaying by 0.99 every 10 epochs until reaching 1e-5. Training is supervised by binary cross-entropy loss for point-wise affordance prediction. Similarly, the low-level planner is trained with Adam (initial learning rate 1e-4, batch size 32) and the same scheduler, decaying to 1e-6.

\noindent\textbf{Inference Protocol.} During inference, we use Grounding-DINO~\cite{liu2023grounding} to detect objects and obtain bounding boxes, annotating each with its ID. The directional text is drawn from our predefined relation set $\mathcal{D}$, comprising nine spatial relationships: \textit{left}, \textit{right}, \textit{front}, \textit{behind}, \textit{left-front}, \textit{left-behind}, \textit{right-front}, \textit{right-behind}, and \textit{on}. For the mid-level spatial mapper, we cluster affordance maps $\mathcal{H}$ with K-means ($k=2$) to form a coarse compositional affordance map, which conditions the low-level diffusion planner. In the low-level planner, object bounding boxes are used to build a mask for computing a truncated signed distance function (TSDF) map that includes detected objects. The TSDF map and the point cloud of the object to be placed guide collision avoidance. To construct the fine compositional affordance map for guidance, we first concatenate multiple predicted affordance maps, compute their mean, and combine it with the original predictions. Then, we apply a max operation across all maps to obtain a fine-grained affordance prediction. The resulting fine-grained affordance map is then used to guide object placement toward regions of high affordance that align with the specified language instruction. For test-time guidance $\mathcal{J}$, we set $\lambda_\text{a} = 500$ and $\lambda_\text{c} = 1000$.

\noindent \textbf{Inference Time.} The MLLM requires $3.498 \pm 1.351$\,s, while the affordance mapper and diffusion planner achieve much lower latencies of $0.102 \pm 0.037$\,s and $0.165 \pm 0.009$\,s, respectively. Note that the MLLM's inference time is influenced by local network conditions. Both lower-level models are evaluated on one NVIDIA RTX~4500 Ada GPU.

\section{Experiments}
\label{sec:experiments}
This section demonstrates that \methodname: (1) outperforms representative baselines, (2) benefits from its hierarchical design with test-time guidance, and (3) can be directly deployed on a real robot (Hello Robot Stretch 3) for diverse human-oriented placement tasks.

\subsection{Experimental Setups}

\vspace{-4pt}\noindent \textbf{Benchmarks.} 
Three evaluation splits are prepared: (1) \textbf{Syn.\ Easy}: Synthetic scenes consisting of 5 to 8 objects placed sparsely yet neatly. (2) \textbf{Syn.\ Hard}: Synthetic scenes consisting of 8 to 12 objects densely placed so that physical plausibility has to be fully respected.  (3) \textbf{Real}: Real-world scenes collected across environments to assess generalization.


\noindent \textbf{Evaluation Protocols.} We adopt metrics suggested by previous works \cite{ramrakhya2024seeing, yuan2024robopoint, mo2021o2oafford} shown to be effective: (1) \textbf{Placement Accuracy (PA)} (\%): the object is placed within an annotated region that is semantically aligned with the instruction input.
(2) \textbf{Physical Plausibility (PP)} (\%): the placed object does not collide with other objects, except for the receptacle object (e.g., a table). A placement is collision-free (100\%) if the object avoids all collisions on the receptacle; otherwise, it is 0\%. (3) \textbf{Success Rate (SR)} (\%): a placement is considered successful if it satisfies the above two criteria. 
 
\noindent \textbf{Baselines.} 
We benchmark against several representative approaches. LLaVA \cite{liu2023llava, li2024llavaonevisioneasyvisualtask} and GPT-4o \cite{gpt4o} are state-of-the-art MLLMs. Following ~\cite{ramrakhya2024seeing, yuan2024robopoint}, we prompt these models to output a placement bounding box, from which five candidate points are randomly selected as the final output. RoboPoint~\cite{yuan2024robopoint} is an instruction-tuned MLLM tailored for robotic tasks such as placement and navigation. We also include Semantic Placement (SP-Policy)~\cite{ramrakhya2024seeing}, a model capturing visual common sense, retrained on our dataset for fairness. 
To ensure fairness, we use the expected post-placement bounding box as a proxy for object geometry when constructing prompts~\cite{feng2023layoutgpt}, since encoding 3D spatial constraints into MLLMs is challenging.
 
\begin{figure*}
    \vspace{0.1 in}
    \centering
    \includegraphics[width=\linewidth]{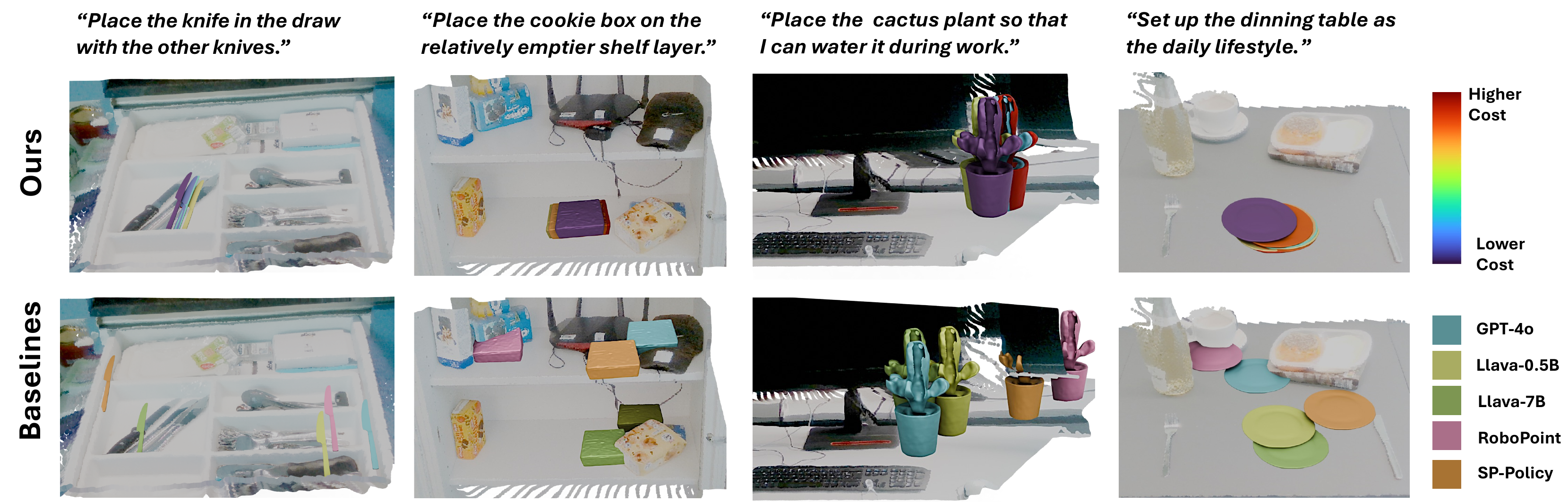}
    \vspace{-0.2 in}
    \caption{Predicted placement by ours and the baselines. Our method not only captures the preference imposed by the instruction, but respects spatial constraints as well.}
    \label{fig:qualitative}
    \vspace{-0.25 in}
\end{figure*}
\subsection{Results and Discussions}
\vspace{-4pt}As shown in Tab.~\ref{tab:benchmark}, our method outperforms all baselines on all metrics in all test splits.
Regarding placement accuracy, SP-Policy \cite{ramrakhya2024seeing} performs worst, with an average accuracy of 27.40\% on synthetic splits, likely due to its limited language understanding. In contrast, MLLMs such as GPT-4o \cite{gpt4o} exhibit much stronger reasoning capabilities, significantly improving the accuracy by an average of 38.60\% on synthetic splits. 
Nonetheless, besides positioning accuracy, all baseline methods struggle to generate physically plausible placement configurations, as they fail to incorporate spatial constraints imposed by the 3D scene and geometric properties of the objects to be placed. This limitation results in a deterioration in physical feasibility, especially in more cluttered scenes, with performance dramatically dropping by ca.\ 30\% compared to ours in Syn.\ Hard split, ultimately reducing the success rate.
In contrast, our method bridges the gap between MLLMs and spatial models by leveraging both the reasoning ability of MLLMs and the geometric-aware planning capacity of spatial reasoning models within a hierarchical framework. Further enhanced by test-time guidance, our approach significantly outperforms all baselines across all metrics, leading to an average of 26.94\% higher success placement than the runner-up method on synthetic splits. 
Notably, our model exhibits strong generalization in real-world scenarios, achieving a success rate of 70.33\% on the Real split, despite being primarily trained on cost-effective synthetic data. Fig.~\ref{fig:qualitative} showcases multiple real-world test results, highlighting the effectiveness of our approach.






\begin{table}[t]
    \centering
    \small
    \setlength{\tabcolsep}{12.5pt}
    \fontsize{7}{8}\selectfont
    \begin{tabular}{@{}r|ccc@{}}
        \toprule
        \textbf{Method} & \textbf{PA} $\uparrow$ & \textbf{PP} $\uparrow$ & \textbf{SR} $\uparrow$ \\
        \cmidrule(lr){1-1}\cmidrule(lr){2-4}
        \textbf{Ours [Full Model]}         & \textbf{71.33}  & \textbf{81.52}   & \textbf{63.16} \\
        \cmidrule(lr){1-4} 
        w/o High-Level MLLM \textbf{[V1]}  & 41.18  &  75.73   & 31.61 \\
        w/o Mid-Level Mapper \textbf{[V2]}      & 48.53  &  71.73   & 34.67\\
        w/o Low-Level Planner \textbf{[V3]}                & 60.47  &  59.93   & 43.20\\
        w/o Test-Time Guide.  \textbf{[V4]}            & 64.80  &  67.47   & 46.67  \\
        \cmidrule(lr){1-4} 
        trained on original data \textbf{[D1]} &55.40 & 79.21 & 46.09\\
        trained on ClutterGen \cite{jia2024cluttergen} \textbf{[D2]} &42.22 & 73.30 &37.17\\
        \bottomrule
    \end{tabular}
    \caption{Ablation results on \textbf{Syn. Hard}. \textbf{PA}: \underline{P}lacement \underline{A}ccuracy. \textbf{PP}: \underline{P}hysical \underline{P}lausibility. \textbf{SR}: \underline{S}uccess \underline{R}ate.}
    \label{tab:ablation}.  
    \vspace{-0.4 in}
\end{table}

\subsection{Ablation Studies}
\vspace{-4pt}We conduct ablation experiments to validate the importance of each component and the effectiveness of our data augmentation. Tab.~\ref{tab:ablation} reports results on the Syn.\ Hard split. \textbf{V1} removes the MLLM, feeding free-form language instead of several structured plans $\mathcal{S}$ to the mid-level mapper and discarding compositional guidance $\mathcal{J}_{\text{afford}}$. \textbf{V2} uses a single structured plan with the diffusion planner, disabling multi-plan composition via $\mathcal{J}_{\text{afford}}$ and affordance maps $\mathcal{H}$. \textbf{V3} removes the diffusion planner, sampling directly from the coarse affordance map $\bar{\bm{H}}$, and \textbf{V4} removes test-time guidance. For fairness, \textbf{V1} and \textbf{V2} are re-trained with the required input format.
To further examine the impact of training data quality, we compare models trained on two data sources: \textbf{D1}, the original dataset without augmentation, and \textbf{D2}, data synthesized using a recent data generator, ClutterGen \cite{jia2024cluttergen}. 

\begin{table}[t]
    \centering
    \small
    \setlength{\tabcolsep}{6.5pt}
    \fontsize{7}{8}\selectfont
    \begin{tabular}{@{}r|ccccc|c@{}}
        \toprule
        \textbf{Method}  & \textbf{Box} & \textbf{Biscuit} & \textbf{Spoon} &  \textbf{Mouse} & \textbf{Keyboard} & \textbf{SR} $\uparrow$  \\
        \cmidrule(lr){1-1}\cmidrule(lr){2-7}
        GPT-4o \cite{gpt4o}        &   7 / 10 & 2 / 10  &2 / 10 &  4 / 10  & 1 / 10 & 32\%\\

        \textbf{Ours}     & \textbf{9} / 10 & \textbf{8} / 10& \textbf{8} / 10 &    \textbf{7} / 10 & \textbf{6} / 10 &  \textbf{76\%} \\ 

        \bottomrule
    \end{tabular}
    \caption{Real-world manipulation success rate of 5 tasks, evaluated on the success rate metric.}
    \label{tab:realmanip}.  
    \vspace{-0.35 in}
\end{table}

\begin{figure*}
    \vspace{0.1in}
    \centering
    \includegraphics[width=\linewidth]{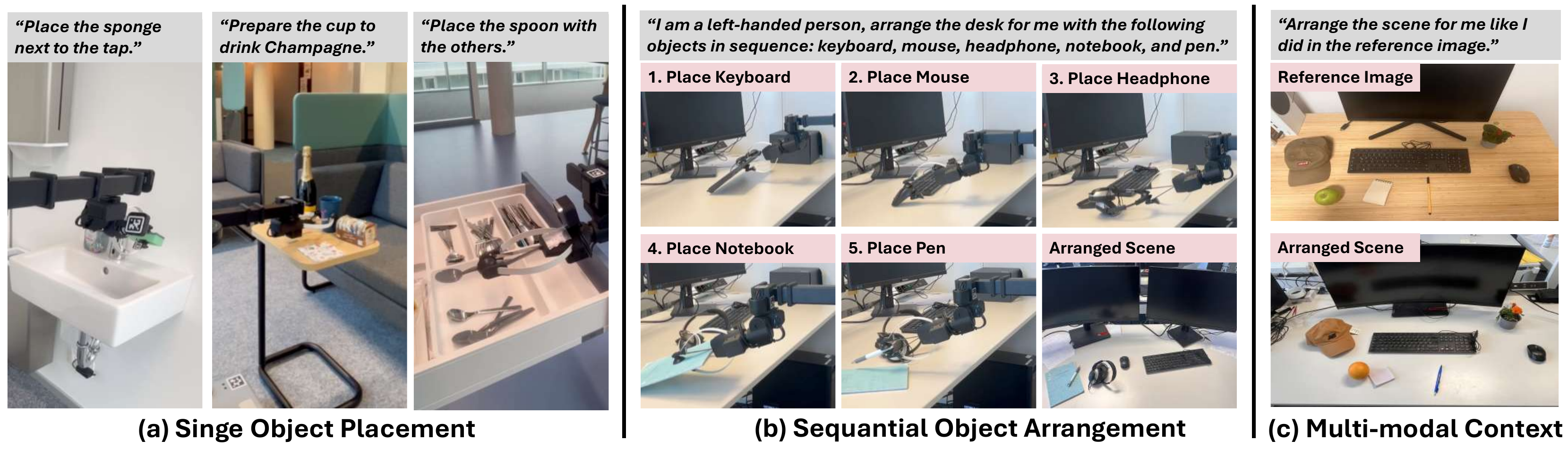}
    \vspace{-0.2 in}
    \caption{Robot integrated with our model can execute versatile downstream arrangement tasks in the real world.}
    \label{fig:downstream}
    \vspace{-0.3 in}
\end{figure*}

\noindent \textbf{Impact of High-Level MLLM Translator (V1).} 
 The 31.55\% drop in success rate underscores the importance of adopting MLLM to acquire structured language plans to ease the semantic reasoning process. Without it, the model struggles to generalize and capture implicit constraints or preferences in free-form language.

\noindent \textbf{Impact of Mid-Level Spatial Mapper (V2).} We observe a 28.49\% drop in success rate, primarily due to inaccurate position estimation. This result underscores the importance of using a spatial mapper to explicitly compose multiple context-aware plans, leading to more robust performance.

\noindent \textbf{Impact of Low-Level Diffusion Planner (V3).} Without the geometric-aware conditioning of the objects to be placed, it significantly reduces physical plausibility by 21.59\%, further lowering the success rate.

\noindent \textbf{Impact of Test-Time Guidance (V4).} We observe a decrease in both placement accuracy and physical plausibility, collectively leading to a 16.49\% drop in the success rate. This highlights the effectiveness of incorporating test-time information, which enforces explicit constraints through analytical cost functions and enhances generalization.

\noindent \textbf{Impact of Data Augmentation (D1).} We observe a 17.07\% decrease in success rate due to the significant performance degradation in placement accuracy. \textbf{D1} suffers from limited data diversity, restricting the model’s generalization.

\noindent \textbf{Impact of Semantically Meaningful Data (D2).} Data produced by ClutterGen~\cite{jia2024cluttergen} lacks semantic awareness, which limits the model's ability to capture meaningful object placement patterns. This deficiency further results in a substantial degradation of the placement accuracy.


\vspace{-4pt}\subsection{Real Robot Experiments}
\vspace{-4pt} 
In real-world experiments, we deploy our model on a mobile manipulator, Hello Robot Stretch 3.

For integration, we first obtain a complete object mesh $\bm{O}$ using a single-view reconstruction pipeline~\cite{long2023wonder3d, muller2022instant}. 
For grasping, we estimate the object pose $\bm{T}_{\mathrm{C}^\prime\mathrm{O}}$ by aligning $\bm{O}$ with the segmented point cloud in an observation frame $\mathrm{C}^\prime$, and apply a grasp detector~\cite{fang2020graspnet} to generate candidate grasps, selecting a final grasp pose $\bm{T}_{\mathrm{C}^\prime\mathrm{E}}$. Given the predicted placement pose $\bm{T}_\mathrm{CO}$ in frame $\mathrm{C}$, the target end-effector pose in the robot base frame $\mathrm{B}$ is computed as
\(
\bm{T}_\mathrm{BE} = \bm{T}_\mathrm{BC}\, \bm{T}_\mathrm{CO}\, (\bm{T}_{\mathrm{C}^\prime\mathrm{O}})^{-1} \bm{T}_{\mathrm{C}^\prime\mathrm{E}},
\)
where $\bm{T}_\mathrm{BC}$ is from the robot’s kinematic tree. The robot uses its motion planner to reach $\bm{T}_\mathrm{BE}$ and release the object.
As shown in Fig.~\ref{fig:downstream}, our placement model can be used to conduct several downstream tasks in various environments, showcasing great generalization and versatility (also see the video for more manipulation results).

\noindent \textbf{Single Object Placement.} We evaluate our model's real-world generalization on novel scenes with unseen objects (e.g., biscuit, spoon, mouse) and environments (e.g., office, restroom, kitchen), using human instructions like \textit{“place the spoon with the others”}. We select 5 tasks to benchmark our method against the runner-up~\cite{gpt4o} tested in simulation. For fairness, we use the same motion planning API for both methods.
As shown in Tab.~\ref{tab:realmanip}, each task is executed for 10 trials, and our model achieves an overall success rate of 76\% across 50 trials, demonstrating strong robustness and generalization. Fig.~\ref{fig:downstream} (a) shows more qualitative examples. 

\noindent \textbf{Sequential Object Arrangement.} This task requires sequential placement of multiple objects, increasing complexity over single object placement~\cite{mo2021o2oafford, ramrakhya2024seeing}. At each step, the object to be placed is absent, and the scene evolves as new objects are added, further compounding the challenge. We tested our model on two scenarios: \textit{office desk arrangement} and \textit{dining table setup}. As shown for the office desk in Fig.~\ref{fig:downstream}(b), the model adapts to the changing scene state and successfully places objects according to user preferences while maintaining physical feasibility, thanks to the MLLM's reasoning and the diffusion planner's test-time guidance.

\noindent \textbf{Multi-Modal Context.} 
Images can also serve as goal states to capture human preferences and guide model inference, providing richer context than language prompts.
Using the MLLM’s strong visual understanding, our model identifies objects similar to those to be placed and infers structured placement plans, while lower-level spatial reasoning models ensure physically plausible configurations.
As shown in Fig.~\ref{fig:downstream} (c), given a reference image of a home work desk, our model captures the underlying arrangement preferences and transfers them to a new, sparsely populated desktop, placing additional items coherently and contextually.

\section{Conclusion}

In this work, we introduce \methodname, a framework for learning object placement from low-cost synthetic training data. Our hierarchical model leverages the strong reasoning capabilities of MLLMs to capture visual common sense aligned with human preferences, while enabling lower-level spatial reasoning models to produce precise, physically feasible placement actions guided by structured plans from the high-level MLLM.
This design substantially reduces the reliance on extensive human demonstration data by enabling training on diverse synthetic examples augmented from a much smaller set of real-world demonstrations. 
Extensive experiments demonstrate that our model achieves superior performance and robust generalization across a wide variety of scenes, objects, and user instructions. Furthermore, its effectiveness and versatility are demonstrated across several real-world robotic applications, underscoring its potential for assistive tasks in human-suited environments.

\noindent \textbf{Limitations.} Our framework assumes access to object meshes, though recent advances in 3D AIGC~\cite{long2023wonder3d, xu2024instantmesh} and learning-based SfM~\cite{dust3r_cvpr24, mast3r_arxiv24} offer more efficient alternatives to traditional scanning~\cite{ xu2019mid}, requiring only a single image or an unordered collection of images. However, the quality of reconstructed meshes varies and may affect performance. Finally, high-precision placement tasks (e.g., robotic assembly) remain difficult. The issue is shared in several other approaches~\cite{ramrakhya2024seeing, mo2021o2oafford, yuan2024robopoint, liu2024moka, song2025robospatial}. To address this, we suggest future work explore a policy that iteratively refines the object’s pose until the desired placement accuracy is achieved.

\footnotesize{
\bibliographystyle{IEEEtran}
\bibliography{reference}

}

\end{document}